\pdfoutput=1

\documentclass[11pt]{article}

\usepackage{acl}

\usepackage{times}
\usepackage{latexsym}

\usepackage[T1]{fontenc}

\usepackage[utf8]{inputenc}

\usepackage{microtype}
\usepackage{mathrsfs}
\usepackage{amsmath}
\usepackage{amssymb}
\usepackage{graphicx}
\usepackage{algorithm}
\usepackage{algpseudocode}
\usepackage{multirow}
\usepackage{booktabs}
\usepackage{url}
\usepackage{CJKutf8}
\usepackage{adjustbox}
\usepackage{enumitem}
\usepackage{lipsum}
\setlist{topsep=1pt,itemsep=1pt,partopsep=1pt, parsep=1pt}
\title{SNP2Vec: Scalable Self-Supervised Pre-Training for \\Genome-Wide Association Study}

\author{Samuel Cahyawijaya$^1$\thanks{\hspace{1mm} These authors contributed equally.}, Tiezheng Yu$^{1*}$, Zihan Liu$^{1*}$, \textbf{Tiffany T.W. MAK$^{2,3}$}, \\
\vspace{6pt} \textbf{Xiaopu Zhou$^{2,3,4}$}, \textbf{Nancy Y. Ip$^{2,3,4}$}, \textbf{Pascale Fung$^1$} \\
$^1$Center for Artificial Intelligence Research (CAiRE), Department of Electronic and Computer Engineering, \\
The Hong Kong University of Science and Technology, Hong Kong, China \\ 
\vspace{6pt} \texttt{\{scahyawijaya,tyuah,zliucr,pascale\}@ust.hk}\\
$^2$Division of Life Science, State Key Laboratory of Molecular Neuroscience, Molecular Neuroscience Center, \\
The Hong Kong University of Science and Technology, Clear Water Bay, Kowloon, Hong Kong, China \\
\vspace{6pt} \texttt{\{tiffanytze,xpzhou,boip\}@ust.hk}\\
$^3$Hong Kong Center for Neurodegenerative Diseases, Hong Kong Science Park, Hong Kong, China \\
\vspace{6pt} \texttt{\{tiffanytze,xpzhou,boip\}@ust.hk}\\
$^4$Guangdong Provincial Key Laboratory of Brain Science, Disease and Drug Development; \\
Shenzhen–Hong Kong Institute of Brain Science, HKUST Shenzhen Research Institute, Shenzhen, China \\
\texttt{\{xpzhou,boip\}@ust.hk}\\\
}

\begin{document}
\maketitle
\begin{abstract}

Self-supervised pre-training methods have brought remarkable breakthroughs in the understanding of text, image, and speech.
Recent developments in genomics has also adopted these pre-training methods for genome understanding.
However, they focus only on understanding haploid sequences, which hinders their applicability towards understanding genetic variations, also known as single nucleotide polymorphisms (SNPs), which is crucial for genome-wide association study.
In this paper, we introduce SNP2Vec, a scalable self-supervised  pre-training approach for understanding SNP. We apply SNP2Vec to perform long-sequence genomics modeling, and we evaluate the effectiveness of our approach on predicting Alzheimer's disease risk in a Chinese cohort. Our approach significantly outperforms existing polygenic risk score methods and all other baselines, including the model that is trained entirely with haploid sequences. We release our code and dataset on \url{https://github.com/HLTCHKUST/snp2vec}.

\end{abstract}

\section{Introduction}

Self-supervised pre-training has become an indispensable step for almost all natural language processing (NLP) tasks~\cite{devlin2019bert,liu2019roberta,yang2019xlnet}. Pre-trained language models, thanks to the usage of massive text corpora, are effective in handling data scarcity and generalizing to unseen examples~\cite{brown2020gpt3,cahyawijaya2021indonlg,wilie2020indonlu,yu2021adaptsum,liu2021crossner,winata2021-multilingual}.
Inspired by the success of pre-trained language models, pre-trained genomic models have been proposed to cope with genomic sequence prediction tasks~\cite{zaheer2020big,ji2021dnabert}. However, these models only focus on modeling the four nucleobases (i.e., A, T, C, and G), while ignoring genomic variations in the pre-training stage. Although they are effective in haploid pattern analysis, such as promoter region and chromatin-profile prediction, they fail to tackle more complex and challenging tasks, such as genome-wide association study (GWAS)~\cite{wtccc2007gwas,corvin2010gwas,bush2012gwas}, which require an in-depth understanding of long genomic sequences and the genomic variation between a homologous chromosome pair.

To address these shortcomings, we introduce a self-supervised pre-training approach called SNP2Vec, which leverages the single-nucleotide polymorphism (SNP, pronounced `snip') information gathered from a large-scale SNP database to inject genomic variations in the pre-training stage. SNP2Vec enables the model to learn the semantics of a diploid sequence (genotype) pattern in a diploid cell. We apply SNP2Vec to a linear-attention model, Linformer~\cite{wang2020linformer}, to allow the model to encode long genomic sequences for Alzheimer's disease risk prediction in a Chinese cohort. We compare SNP2Vec with non-pretrained models, as well as an existing strong baseline polygenic risk scoring (PRS) model, to demonstrate the effectiveness of our approach.

Our contributions are summarized as follows:
\begin{itemize}
    \item We are the first to introduce a scalable self-supervised pre-training approach (SNP2Vec) to learn genomic variations, which is popular for genome-wide association study.
    \item We demonstrate a method for modeling long diploid sequences with a length of >20,000 base pairs (bps) using an attention-based model within a single forward pass.
    \item We demonstrate the effectiveness of SNP2Vec, which significantly outperforms all the baselines, including a widely-used polygenic risk scoring (PRS) method, by 5-7\% accuracy and AUROC for the Alzheimer's disease prediction task in a Chinese elderly cohort.
    \item We conduct comprehensive analyses to show the effectiveness of SNP encoding and Byte Pair Encoding (BPE) tokenization compared to the other commonly used methods for genomics modeling.
\end{itemize}

\section{Related Works}
\label{sec:related_works}


\subsection{Genome-Wide Association Study}

To this day, predicting the risk of hereditary diseases from a given genotype is done through genome-wide Associaction Study (GWAS) by applying a polygenic risk score (PRS). PRS utilizes GWAS data to identify important single nucleotide variations (SNVs) over a certain range from the gene of interest. The SNVs are first filtered according to a statistical measure to reduce the bias towards a certain population and the filtered SNVs are then used to build a classifier, which can be applied to a new genotype to determine the likelihood of getting the disease.
This method has been applied by many works and has provided valuable insights for researchers to diseases including heart attack, diabetes, and different types of cancer~\cite{lello2019genomic}.
Moreover, PRS model has also been used in research and clinical practice for Alzheimer’s disease \cite{zhou2021polygenic}. 
Nevertheless, all these methods fail to incorporate the patterns of the genomics sequence that determines the actual function. This is likely to lead the model towards non-representative bias, especially when the experimental data is small.

\subsection{Statistical Modeling for Genomics}


\paragraph{Tokenization in Genomics}
$k$-mer (synonymous to n-gram) tokenization is the most commonly used tokenization method in existing genome modelling works \cite{min2017kmeremb,shen2018recurrent}. 
Gapped $k$-mer tokenization \cite{ghandi2014enhanced,shrikumar2019gkmexplain} is  a more efficient variant of $k$-mer tokenization by introducing the gap parameter $L$, which constitutes the stride between each $k$-mer window. However, the gapped $k$-mer approach will lead to the loss of some information when $L$ is larger than $k$. In recent years, subword tokenization approaches~\cite{sennrich2015neural,kudo2018sentencepiece} have also been explored in genomics~\cite{zaheer2020big}.

\paragraph{Machine Learning in Genomics}
The support vector machine (SVM) is a traditional machine learning approach used to quickly and accurately interpret the nonlinear gapped k-mer \cite{shrikumar2019gkmexplain}. \citet{hill2018deep} leverage a deep recurrent neural network (RNN) to discover complex biological rules to decipher RNA protein-coding potential. 
\citet{zhuang2019simple} incorporate convolutional neural network (CNN) to predict enhancer–promoter interactions with DNA sequence data. 
\citet{shen2018recurrent} introduce a RNN to predict transcription factor binding sites. 
They treat each k-mer as a word and pre-train a word representation model though word2vec algorithm~\cite{mikolov2013efficient}. \citet{zaheer2020big} propose BigBird and pre-train it on the human reference genome
and improves the performance on downstream tasks. 

\subsection{Self-Supervised Pre-training}

Recently, using self-supervised pre-training models on large scale unlabeled data and then fine-tuning them using a small amount of labeled data has become the norm in machine learning. 
BERT \cite{devlin2019bert} is a deep bidirectional transformer pre-trained on BooksCorpus \cite{zhu2015aligning} (800M words) and English Wikipedia (2500M words) for language understanding. \citet{liu2019roberta} introduces Roberta, which has a similar architecture as BERT but trained on a much larger corpus (160GB of text) and consequently achieves better performance. In recent years, pre-training generative models \cite{radford2019language,raffel2019exploring,lewis2019bart} has significantly improved the performance of various language generation tasks such as machine translation, question answering, conversational AI, etc. 

Self-supervised learning approaches have also been adopted in genomics~\cite{zaheer2020big,ji2021dnabert} and proteomics~\cite{madani2020progen,elnaggar2020prottrans}.
These methods pre-train models using large-scale unlabelled datasets such as the human reference genome from the Genome Reference Consortium (GRC)~\cite{church2011hg19,schneider2017refgen38} and protein sequence databases such as SWISS-PROT and TrEMBL~\cite{boeckmann2003swissprot}. In this paper, we focus on genomics and conduct the human reference genome for pre-training. Genomics data does not have the same structure as human languages; it has  no known syntax or grammatical rules and it consists of very long sequences with only a number of differences between each human subject. 

\begin{figure*}
    \centering
    \resizebox{1.0\linewidth}{!}{
    \includegraphics{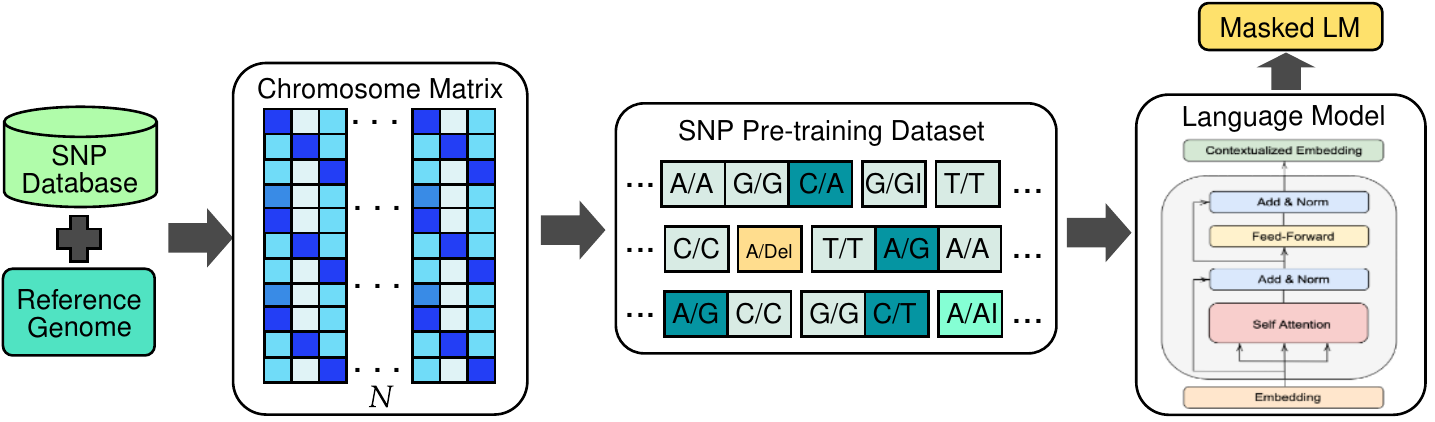}
    }
    \caption{SNP2Vec Pre-training Pipeline. SNP2Vec merges information from reference genome and SNP database to form a chromosome matrix which is then utilized to construst SNP pre-training dataset following the SNP encoding's token format. This pre-training dataset is employed to train a genome language model through the masked language modeling task.}
    \label{fig:snp2vec_pretraining}
\end{figure*}

\section{SNP2Vec}

Existing pre-training methods in genomics, such as BigBird~\cite{zaheer2020big} and DNABERT~\cite{ji2021dnabert}, are only optimized to understand the pattern of a haploid sequence (haplotype) based on the reference genome. This hinders the model from learning genomic variations, which is essential for understanding traits in humans. In contrast to prior works in genomics pre-training, we develop SNP2Vec to enable pre-training for encoding and understanding patterns of genomic variations in a diploid sequence. Figure~\ref{fig:snp2vec_pretraining} depicts the overall structure of the SNP2Vec pre-training method. We elaborate on our SNP2Vec method in 3 subsections: 1) SNP Encoding, i.e., how we encode a diploid sequence as a sequence of SNP tokens; 2) Self-Supervised SNP Dataset, i.e., how we construct a self-supervised dataset using the SNP token; 3) Self-supervised SNP Pre-training, i.e., how we perform self-supervised pre-training for learning the sequence pattern of SNP tokens.

\subsection{Preliminaries}

\begin{figure}[!t]
    \centering
    \resizebox{1.0\linewidth}{!}{
    \includegraphics{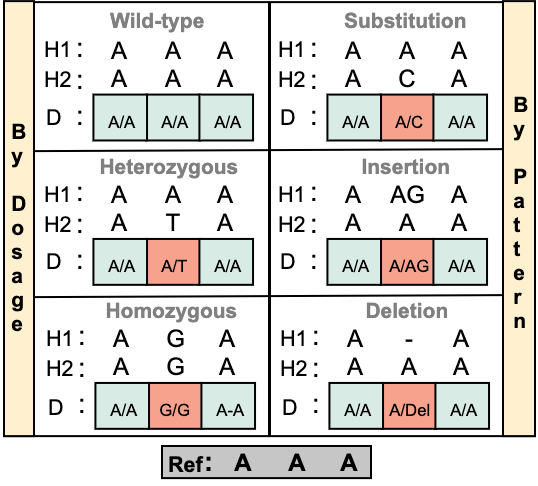}
    }
    \caption{Diploid sequence variations. The box on the top-left shows the wild-type sequence, while others are its variations. \textbf{H1} and \textbf{H2} denote the haploid sequence for each parent allele. \textbf{D} represents the diploid sequence of the two alleles.}
    \label{fig:haplo_diplo}
\end{figure}

\paragraph{What are haploid and diploid sequences?} 
A diploid is a cell or organism that has paired chromosomes, one from each parent~\footnote{\url{https://www.genome.gov/genetics-glossary/Diploid}}. Human cells are mostly diploid, except for the sex cells. In this sense, a diploid sequence (genotype) refers to a pair of homologous sequences (allele) inside the diploid chromosome, while a haploid sequence (haplotype) refers to the DNA sequence from the specific allele of the diploid sequence. The haploid sequence is suitable for understanding the regulatory function of a DNA pattern~\cite{zhou2015dsea,ouyang2008haplotype}, such as determining a binding site for a certain type of protein, as it provides the representation of the actual nucleotides. A diploid sequence, on the other hand, is more suitable for understanding the phenotype~\cite{levy2007diploid,wang2008diploid} over population since it allows understanding of the genomic variations between two homologous DNA sequences, which tells the dosage information and the gene expression level of a variation. These genomic variations are gathered by comparing them to a genome reference sequence, and they can be categorized based on its dosage, i.e., wild-type (normal), heterozygous, or homozygous, and based on their differences, i.e., substitution, insertion, and deletion. 
The depiction of haploids and diploids along with their variations is shown in Figure~\ref{fig:haplo_diplo}.

\paragraph{How do we get the haploid and diploid sequence?} As most human cells are predominantly diploid, performing genome sequencing on such homologous chromosome pair will produce a diploid sequence rather than a haploid sequence, because the primer binds to both of the homologous regions from each chromosome~\cite{ye2012primer}. Extracting haploid sequences from a diploid sequence requires an additional step through an estimation process called phasing~\cite{stephens2001phasing}. Despite their effectiveness, the quality of phasing methods~\cite{browning2007beagle,browning2009beagle,patterson2014whatshap} is not perfect and tends to decrease significantly especially when the gap between the SNPs is large~\cite{choi2018phasing}.

\subsection{SNP Encoding}
\label{subsec:snp_encoding}

\begin{figure}[!t]
    \centering
    \resizebox{1.00\linewidth}{!}{
    \includegraphics{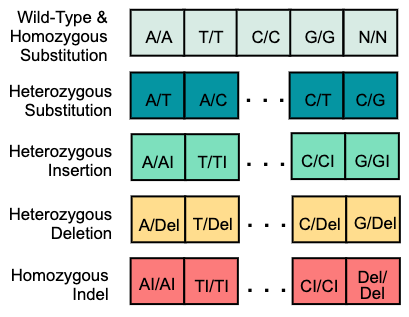}
    }
    \caption{SNP tokens consist of a total of 66 types of token covering all possible variations in a diploid sequence including wild-types, heterozygous variations, and homozygous variations.}
    \label{fig:snp_tokens}
\end{figure}

We first extend the nucleotide tokens from 5 token types {`A', `T', `C', `G', and `N'} into 11 tokens by adding 6 insertion-deletion (indel) tokens {`AI', `TI', `CI', `GI', `NI', and `DEL'}, where `XI' token represents any insertion after the nucleotide `X', and `DEL' represents the nucleotide deletion. There can be many different possibility for insertion, e.g., a nucleotide `T' can be inserted into ``TG'', ``TGGG'', or ``TAAA''; therefore we aggregate all the insertions into a single token to reduce the sparsity of the indel representaton as indel occurs relatively rarely compared to substitution, with an around 1:5 ratio~\cite{chen2009variation}. To encode a diploid sequence, we construct all the combinations with replacement ($_nC^R_r$) of the 11 haploid and indel tokens  with $n=11$ and $r=2$, producing a total of 66 types of SNP tokens consisting of wild-type, heterozygous, and homozygous variation tokens. The resulting SNP tokens are represented as `X$_1$/X$_2$'', where `X$_1$' and `X$_2$' denote aligned nucleotide or indel tokens from the two alleles ordered alphabetically. A depiction of the SNP tokens is shown in Figure~\ref{fig:snp_tokens}. To reduce the size and facilitate more straightforward representation for downstream processes such as pre-processing, tokenization, and modeling, we map the SNP tokens into a single character representation. The mapping of the SNP token into a single character representation is shown in Appendix~\ref{app:snp-tokens-mapping}.

By incorporating the SNP encoding, variant calling information gathered from the DNA sequencing machine can be directly converted into a sequence of SNP tokens, that are then used for the model fine-tuning and inference. However, this is not directly applicable for self-supervised pre-training since DNA sequencing data is hard to obtain and it is unethical to share publicly as it contains very sensitive and personal information of the human subject. In the next section, we discuss in detail how we can construct an inexpensive and reliable pre-training dataset to perform self-supervised pre-training on the SNP tokens by utilizing publicly available genomics data sources.

\subsection{Self-Supervised SNP Dataset}

Prior self-supervised pre-training approaches in genomics~\cite{zaheer2020big,ji2021dnabert} only utilize the human reference genome~\cite{church2011hg19,schneider2017refgen38} as the unlabelled data for haploid genomics pre-training, the latter does not capture any genomic variations. We extend these haploid modeling techniques into a diploid modeling method, which allows the model to learn patterns of genomic variations by generating unlabelled pre-training data for learning SNP tokens. More specifically, we use the genome sequence from the human reference genome and genome variation from a large-scale SNP database, namely dbSNP~\cite{smigielski2000dbsnp}, to generate the pre-training data.


\paragraph{Human reference genome} The human reference genome is a genome sequence derived from the DNA collected from a number of people~\cite{pollard2017chromosome}, which was first released in 2000
and is periodically updated. There are two most commonly used versions of the human reference genome, namely GRCh37~\cite{church2011hg19}~\footnote{\url{https://www.ncbi.nlm.nih.gov/assembly/GCF_000001405.13/}} and GRCh38~\cite{schneider2017refgen38}~\footnote{\url{https://www.ncbi.nlm.nih.gov/assembly/GCF_000001405.26/}}. 
A human reference genome consists of the genome sequence information for all human chromosomes with $\sim$3B sequence length in total. Most of the positions are mapped and represented as either `A', `T', `C', or `G', while the others are unmapped and flagged with the unknown (`N') token.

\paragraph{dbSNP} dbSNP~\cite{smigielski2000dbsnp}~\footnote{\url{https://www.ncbi.nlm.nih.gov/projects/SNP/snp_summary.cgi}} is a central public repository of human SNPs. dbSNP covers a broad collection of simple genetic variations with a length of variation $\leq$50 bps long, which includes single-base nucleotide substitutions, small-scale multi-base deletions, and small-scale multi-base insertions. 
A single SNP in the dbSNP contains the following information: chromosome number, position in the chromosome, SNP identifier, reference sequence (\texttt{REF}), alternative sequence(s) (\texttt{ALTS}), probability of the \texttt{REF} and \texttt{ALTS}, and other metadata. The \texttt{REF} is a single-base or multi-base sequence that comes from the human reference genome used for detecting the SNPs. The \texttt{ALTS} can consist of one or more alternative variations and each can represent a substitution, a deletion, or an insertion.

\paragraph{Dataset Construction} We construct a pre-training dataset consisting of sequences of SNP tokens by combining the sequence information from the human reference genome and the genomic variations from the dbSNP. For each chromosome, we generate an $11 \times N$ matrix, where $N$ is equal to the length of the corresponding chromosome and $11$ represents the probability of each nucleotide and indel token. We name this matrix a \textbf{chromosome matrix}. We fill the chromosome matrix using all SNPs labelled as \texttt{COMMON} in the dbSNP by filling the corresponding matrix position with the \texttt{REF} and \texttt{ALTS} probability of the corresponding SNP record. 
Since the SNPs from the dbSNP do not cover all of the genome positions, we fill up all the other gap positions with a probability of 1 to the nucleotide token in the corresponding position on the human reference genome.

For constructing the self-supervised pre-training dataset, we closely follow the setup in the typical NLP pre-training dataset construction pipeline. Specifically, we convert the chromosome matrix into a set of segments $S$ where each segment $s \in S$ comprises of a number of SNP tokens. To construct the sentences $S$, we sample multiple sequences from different positions of a chromosome. For each position in the sequence, we apply a sampling function $F$ to collect `X1' and `X2' (the nucleotide or indel tokens on the corresponding position) and construct the SNP token ``X1/X2''. The dataset construction method can be applied to all the chromosome pairs except for the sex chromosome, which is always haploid. The details of our dataset construction approach is shown in Algorithm~\ref{alg:pt-construct}.

\subsection{Self-Supervised SNP Pre-training}

Inspired by BERT~\cite{devlin2019bert}, SNP2Vec is trained using the masked language modeling (MLM) objective using a transformer-based model~\cite{vaswani2017transformer}. The goal of MLM is to predict the representations of the masked tokens given their neighbouring sequence as the context. As complex genomic tasks, such as disease risk prediction, require the understanding long-genome 
sequence (>1000 bps), we apply two methods to process long input sequences. First, we apply a transformer variant with a linear-attention mechanism, which enables the model to reduce the computational complexity from $O(N^2)$ to $O(N)$. Second, we apply a BPE tokenization~\cite{sennrich2015neural} to encode the sequence of SNP tokens to compress the sequence via aggregation of neighbouring tokens. Unlike k-mer~\cite{min2017kmeremb,ji2021dnabert} and gapped k-mer~\cite{ghandi2014enhanced,shrikumar2019gkmexplain} tokenizations, BPE tokenization can merge dynamic-length tokens based on their co-occurrences efficiently without losing any information.

\begin{algorithm}[t]
{\selectfont
\caption{Self-Supervised Pre-training dataset construction for diploid SNP Encoding}
\label{alg:pt-construct}
\textbf{Require:} $C$: chromosome matrix\\
\textbf{Require:} $f$: SNP sampling function \\
\textbf{Require:} $T$: number of iterations\\
\textbf{Require:} $K$: start position threshold\\
\textbf{Require:} $L^{inf}$: lower bound of segment length
\textbf{Require:} $L^{sup}$: upper bound of segment length
\begin{algorithmic}[1]
\State Initialize $S$ = $\emptyset$ 
\State $P$ =  sample $T$ positions from range $[0 \dots K]$
\ForAll{$p \in P$}
  \While{$p < |C|$}
      \State $l \sim \mathcal{U}(L^{inf}, L^{sup})$
      \State $z$ = segment from $p$ to $p + L$ in $C$
      \State $s$ = Sample SNP tokens using $f$ from \phantom . \phantom .\phantom .\phantom .\phantom .\phantom .\phantom .\phantom .\phantom  .\phantom  .\phantom  .\phantom .\phantom  .\phantom  .\phantom  .\phantom  .\phantom  .\phantom  .\phantom  .\phantom  .\phantom .each position in $z$
      \State $S = S \cup s$
      \State $p = p + l$
  \EndWhile
\EndFor
\end{algorithmic}
}
\end{algorithm}

\section{Experiment Settings}
\label{sec:experiment_settings}
\subsection{Dataset}

For building the pre-training data, we utilize GRCh37 as the human reference genome and dbSNP version 153~\footnote{\url{https://ftp.ncbi.nih.gov/snp/archive/b153/00readme.txt}} as the SNP database. We utilize a weighted random sampling based on the probability of SNPs on the corresponding position as the sampling function $f$. For the downstream-task, we construct a dataset of genome sequences for predicting late-onset Alzheimer's disease (LOAD)~\cite{rabinovici2019load} on a Chinese Cohort from 624 Hong Kong elderly with a minimum age of 65. The subjects are diagnosed with Alzheimer's by a medical professional through the Montreal Cognitive Assessment (MoCA) test~\cite{nasreddine2005moca} adjusted for the demographic information. 
Out of 624 subjects, 384 are labelled as Alzheimer's disease carriers (ADs) and 240 are labelled as non-carriers (NCs). For the genome sequence, we collect sequencing data from the APOE region located in chromosome 19 from each subject, which is known to be highly correlated with Alzheimer's disease in the Chinese cohort~\cite{zhou2019apoe,zhou2020adcn}. We use BWA-MEM~\cite{li2013aligning} assembler to align the sequencing data with the human reference genome.

\subsection{Training and Evaluation Setting}
\label{subsec:training_setup}

For our experiment, we build a BPE tokenizer with a vocabulary size of 32,000 tokens. We pre-train a 6-layers linear-attention transformer-based model, Linformer~\cite{wang2020linformer}, using a  maximum sequence length of 4,096 tokens, a sequence projection length $k$ of 128 tokens, and a model dimension size of 512. For simplicity, we refer to our pre-trained SNP2Vec model as \textbf{Dipformer}. The detail hyperparameters of the BPE tokenizer and the Dipformer model are described in Appendix~\ref{app:model_hyperparam}. We run MLM pre-training for 200,000 steps with a 15\% token replacement rate, where we replace with \texttt{[MASK]} 80\% of the time, replace with a random token 10\% of the time, and keep the token as is 10\% of the time. More detail about the pre-training hyperparameter setting is shown in Appendix~\ref{app:pretrain_hyperparam}.

For the fine-tuning, we apply SNP encoding to the sequencing data, apply BPE tokenization, and add a \texttt{[CLS]} token as the prefix of the sequence to gather the sequence representation for predicting the risk of having Alzheimer's disease. We apply fine-tuning for three input sequence length settings, i.e., only APOE gene with 3,611 bps ( \textbf{APOE only}), APOE with additional 5,000 bps upstream and downstream (\textbf{APOE+10k}), and APOE with additional 10,000 bps upstream and downstream (\textbf{APOE + 20k}). For each experiment, we apply 10-fold cross validation to ensure the result is significance. We evaluate the model performance using three evaluation metrics: accuracy, area under the ROC curve (AUROC), and area under the precision-recall curve (AUPRC). More detail about the fine-tuning setup is described in Appendix~\ref{app:finetuning_hyperparam}.

\subsection{Baselines}

To evaluate the effectiveness of the SNP encoding, we build two different deep learning models using haploid token representation. First, we incorporate DeepSEA~\cite{zhou2015dsea}, a CNN-based model develop for short sequence chromatin profiling tasks ($\sim$200-1000 bps), and then we build another Linformer model pre-trained with the human reference genome using haploid tokens, called \textbf{Hapformer}. For the haploid token fine-tuning, we generate the haploid sequence from the aligned sequencing data. We generate the variant calling data with GATK HaplotypeCaller~\cite{mckenna2010gatk,depristo2011gatk} and apply phasing with Beagle~\cite{browning2007beagle,browning2009beagle}. During fine-tuning, we feed each haploid sequence to the model and fuse the representation using a linear transformation.
We also incorporate a logistic regression model from PLINK~\cite{purcell2007plink}, which is a widely used approach for PRS.

\begin{table}[t!]
\centering
\setlength{\tabcolsep}{15pt}
\resizebox{\linewidth}{!}{
\begin{tabular}{lccc}
\toprule
Models    & Acc       & AUROC & AUPRC \\ \midrule
DeepSEA   & 0.591    & 0.579 & 0.703 \\
PLINK PRS & 0.592    & 0.607 & 0.705 \\
Hapformer & 0.572    & 0.615 & 0.715 \\ \midrule
Dipformer & \textbf{0.643} & \textbf{0.673} & \textbf{0.734}\\ \bottomrule
\end{tabular}
}
\caption{Results of our model and baselines. We refer the pre-trained SNP2Vec model as Dipformer.}
\label{tab:score_ours_vs_baseline}
\end{table}

\begin{table*}[t]
\centering
\setlength{\tabcolsep}{12pt}
\resizebox{0.99\textwidth}{!}{
\begin{tabular}{lcccccccc}
\toprule
Tokenization      & 1-mer        & 3-mer        & 5-mer        & gkm (5,6)    & gkm (6,10)   & gkm (6,14)   & gkm (7,14)   & BPE      \\
Avg. Token Length & 500          & 498          & 496          & 83           & 50           & 36           & 36           & 81.19        \\ \midrule
BoW Linear        & \underline{0.499} & \underline{0.698} & 0.817 & \underline{0.771} & \underline{0.759} & \underline{0.749}  & \underline{0.753} & \textbf{0.783}* \\
CNN (DeepSEA)     & 0.890 & 0.903  & 0.898 & 0.808  & \underline{0.764} & \underline{0.749} & \underline{0.751}  & \textbf{0.811}* \\
Transformer       & \underline{0.727} & 0.788 & 0.825 & 0.785 & \underline{0.771}  & \underline{0.761} & \underline{0.762} & \textbf{0.789}* \\ \midrule
Average           & \underline{0.706} & 0.796 & 0.847 & \underline{0.788} & \underline{0.765} & \underline{0.753} & \underline{0.755} & \textbf{0.795}* \\ \bottomrule
\end{tabular}
}
\caption{Comparison of different tokenization methods in genome modeling (numbers denote the accuracy score), where gkm ($k$,$l$) denotes the gapped $k$-mer tokenization with the gap parameter $l$ constituting the stride between each $k$-mer window. * denotes that BPE significantly outperforms the underlined baselines with a p-value < 0.01.}
\label{tab:score_tokenization}
\end{table*}

\section{Results}
The results of our model and baselines are shown in Table~\ref{tab:score_ours_vs_baseline}. 
We find that Dipformer is able to outperform existing strong baselines, such as DeepSEA and PLINK PRS, by a large margin. This confirms the effectiveness of our SNP2Vec pre-training, and the ability of our Dipformer to capture relevant features for AD prediction. 
Interestingly, Hapformer, which leverages large amounts of genomic sequences for pre-training, only performs comparably to DeepSEA and PLINK PRS. Moreover, by learning genomic variations in a diploid sequence during the pre-training, Dipformer significantly outperforms Hapformer with an around 5-7\% improvement in terms of accuracy and AUROC metrics. This shows that simply using an enormous amount of pre-training data might not necessarily improve the AD prediction, and an effective genomics pre-training approach is essential to guarantee full use of the unlabelled genomics data. More detail on our results is shown in Appendix~\ref{app:results_detail}.

\section{Discussion}

\begin{table*}[t]
\begin{adjustbox}{width=\linewidth,totalheight={\textheight},keepaspectratio}
\begin{minipage}[!b]{0.5\linewidth}
        \centering
        \includegraphics[width=\linewidth]{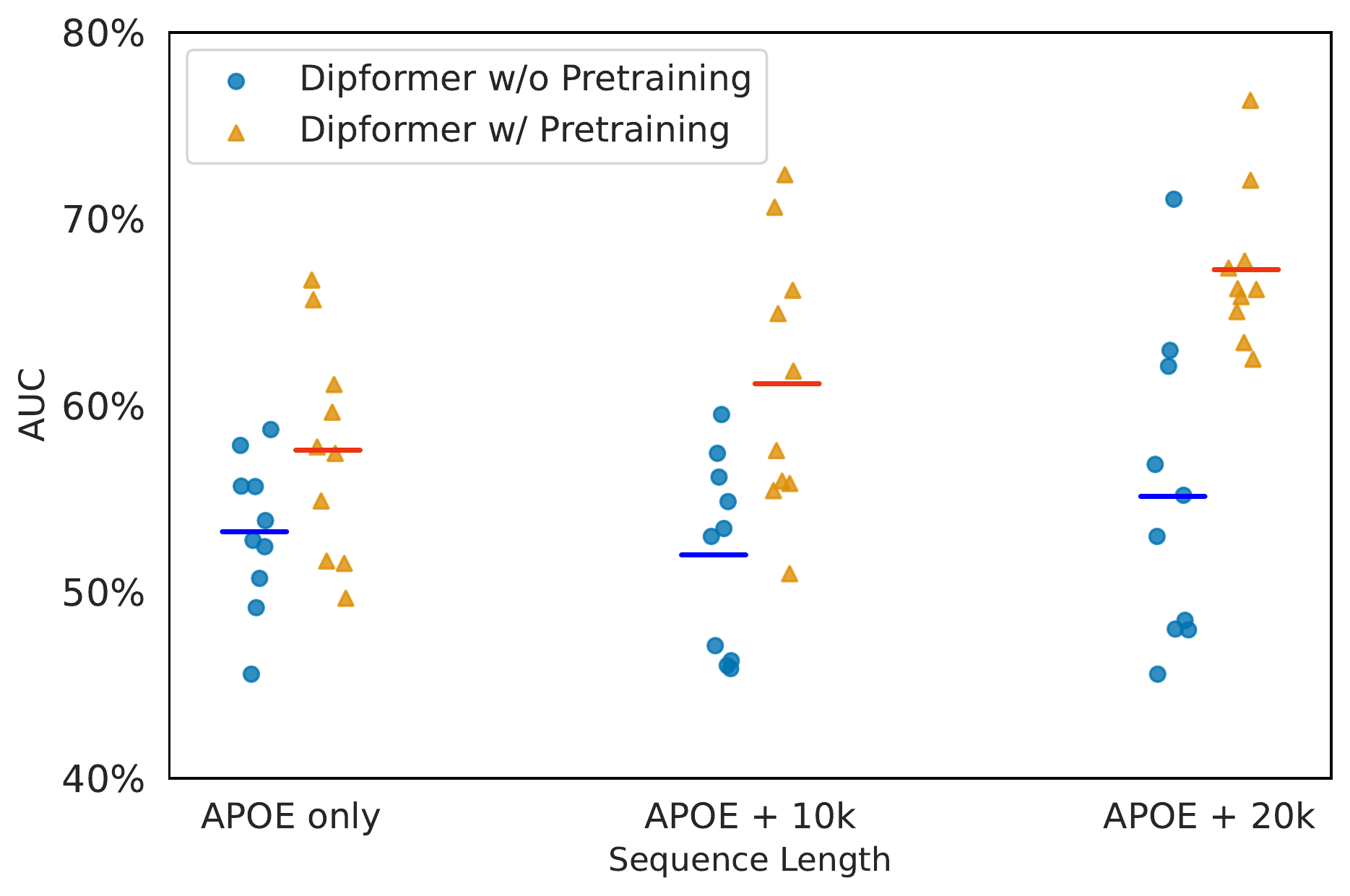}
        \captionof{figure}{10-folds AUROC Performance of Dipformer with and without pre-training on the Alzheimer's disease risk prediction over different sequence length input.}
        \label{fig:effect_of_pre-training}
\end{minipage}
\hspace{6mm}
\begin{minipage}[!b]{0.5\linewidth}
        \centering
        \includegraphics[width=\linewidth]{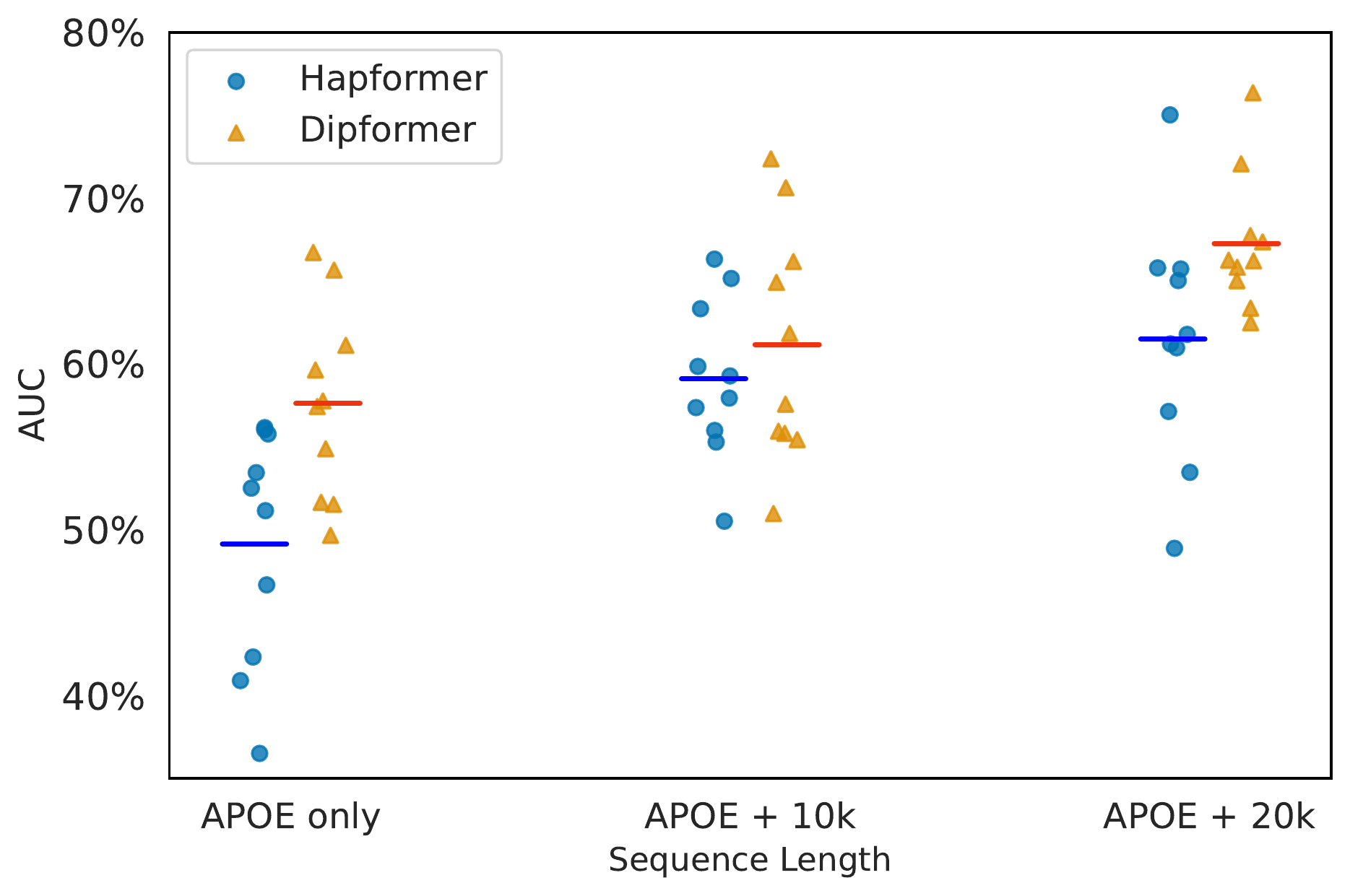}
        \captionof{figure}{10-folds AUROC performance of pre-trained Hapformer and Dipformer on the Alzheimer's disease risk prediction over different sequence length input.}
        \label{fig:effect_of_diplotype}
\end{minipage}
\end{adjustbox}
\end{table*}

\subsection{Effect of Different Tokenization Methods in Genomics}
In this section, we study different tokenization methods for genome modeling, and explore their effectiveness in terms of capturing genomic patterns and features.
We compare BPE tokenization with other common methods, such as $k$-mer and gapped $k$-mer (gkm) with various gap parameters.
To achive this, we conduct experiments on the chromatin profiling dataset from DeepSEA, which consists of 4,863,024 chromatin profiles (4,400,000 training, 8000 validation, and 455,024 test) with 919 labels (690 transcription factor (TF) binding sites, 125 DNase marks, and 104 Histone marks).
Three different models are incorporated in this experiment: a linear model with bag-of-word (BoW) representation, a CNN-based model following the DeepSEA architecture, and a transformer model. 
The models need to predict the TF, DNase, and Histone labels based on the input sequences using various tokenization methods. Hence, for the same model, a more effective tokenization method will lead to a higher prediction accuracy. Additionally, we use the average length of the tokenized sequences to measure the efficiency of different tokenization methods as it determines the input size for the model.

Table~\ref{tab:score_tokenization} provides the effectiveness and averaged token length of different tokenization methods in genome modeling. We find that, on the Linear BoW model, BPE significantly outperforms all other methods except 5-mer. On the CNN model, BPE remarkably surpasses all gapped k-mer methods except for the gkm (5,6). On the Transformer model, BPE performs similarly to 3-mer and gkm (5,6), and significantly outperforms 1-mer and other gkm methods. Moreover, in terms of the averaged score across all three models, BPE performs comparably well to 3-mer, and remarkably outperforms 1-mer and all gkm methods.

\begin{figure}[t]
    \centering
    \includegraphics[width=1.0\linewidth]{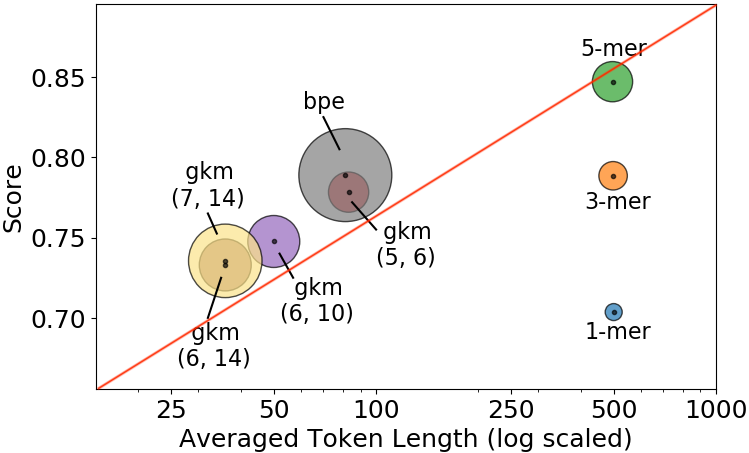}
    \caption{Performance efficiency trade-off of using different tokenization approaches. The score is averaged over the three models (Linear, CNN, and Transformer). The size of the dots represents the vocabulary size of the tokenization method.}
    \label{fig:efficiency_vs_performance}
\end{figure}

Figure~\ref{fig:efficiency_vs_performance} illustrates the trade-off between the performance and efficiency of different tokenization methods. We can see that compared to $k$-mer methods, BPE performs comparably to 3-mer and slightly worse than 5-mer, but it is much more efficient due to a much shorter average length. In addition, BPE remarkably outperforms gkm methods with comparable or slightly worse efficiency. Furthermore, from the size of the dots, we can see that BPE has a much larger vocabulary size compared to other methods, which indicates that BPE can potentially capture richer genomics patterns.

\subsection{Effect of Pre-training for Disease Risk Prediction}
\label{sec:effectiveness_of_pre-training}


In this section, we focus on exploring the effectiveness of pre-training for disease risk prediction. 
Figure~\ref{fig:effect_of_pre-training} illustrates the 10-fold AUROC results of our Dipformer model with and without pre-training on Alzheimer's disease risk prediction. 
The dashes in the figure represent the average AUROC for all 10-fold results. As shown in Figure~\ref{fig:effect_of_pre-training}, the average AUROC scores for pre-trained Dipformer significantly outperform the Dipformer without pre-training in all sequence length settings, APOE + 10k, and APOE + 20.
Table~\ref{tab:effect_of_pre-training} presents the quantitative results with additional metrics. The accuracy, AUROC, and AUPRC scores of pre-trained Dipformer consistently outperform the non-pre-trained Dipformer in all sequence length settings. By increasing the sequence length, the non-pre-trained Dipformer performs slightly better, while the pre-trained Dipformer improves by a large margin. This shows the importance of pre-training for understanding long-sequence features.

\subsection{Effect of the SNP Encoding in Genomics}
To study the effect of the SNP encoding, we pre-train and fine-tune a model with the same genomics data but using haploid tokens called Hapformer, as mentioned in the Section~\ref{sec:experiment_settings}. Figure~\ref{fig:effect_of_diplotype} shows the 10-fold AUROC results of pre-trained Hapformer and Dipformer on the AD risk prediction over different sequence length inputs.  Among all three sequence length settings, Dipformer achieves better average AUROC scores than Hapformer with a p-value of 0.046 for the APOE + 20k setting, which indicates that the improvement of SNP encoding is significant. Meanwhile, the results in Table~\ref{tab:effect_of_pre-training} shows that Dipformer also surpasses Hapformer in all other evaluation metrics. In addition, we also observe that both Hapformer and Dipformer achieve better results when the input sequence is longer. This shows that employing long sequence is essential for handling complex genomics tasks such as disease risk prediction.

\begin{table}[t]
\begin{adjustbox}{width=\linewidth,totalheight={\textheight},keepaspectratio}
\begin{tabular}{lccc}
\toprule
\textbf{Model}          & \textbf{Accuracy} & \textbf{AUROC} & \textbf{AUPRC}   \\ \midrule
\qquad \textit{Without Pre-training} \\ \midrule
Dipformer (APOE only)   & 0.567 & 0.532 & 0.667 \\
Dipformer (APOE + 10k)  & 0.571	& 0.520 & 0.608 \\
Dipformer (APOE + 20k)  & 0.588	& 0.551 & 0.668 \\ \midrule
\midrule
\qquad  \textit{With Pre-training} \\ \midrule
Hapformer (APOE only)   & 0.524 & 0.491 & 0.623 \\
Hapformer (APOE + 10k)  & 0.565	& 0.591 & 0.705 \\
Hapformer (APOE + 20k)  & 0.572	& 0.615 & 0.715 \\ \midrule
Dipformer (APOE only)   & 0.611 & 0.576 & 0.687 \\
Dipformer (APOE + 10k)  & 0.574	& 0.612 & 0.710\\
Dipformer (APOE + 20k)  & \textbf{0.643} & \textbf{0.673} & \textbf{0.734} \\ \bottomrule
\end{tabular}
\end{adjustbox}
\caption{Performance of Dipformer and Hapformer on the Alzheimer's disease risk prediction over different lengths of the input sequences.}
\label{tab:effect_of_pre-training}
\end{table}

\section{Conclusion}

In this paper, we introduce SNP2Vec, a self-supervised pre-training method for understanding genomic variations in a diploid sequence. Unlike prior methods in genomics, SNP2Vec represents each genomics position with a SNP token which allows the model to capture genomic variations which is suitable for understanding complex genomics prediction tasks such as predicting phenotype. By utilizing SNP2Vec, we pre-train a Linformer model called Dipformer and evaluate it for predicting late-onset Alzheimer's disease risk in a Chinese cohort. Experimental results suggest that Dipformer significantly improves the prediction quality by 5-7\% Accuracy and AUROC over all other baselines including the widely used polygenic risk score model from PLINK, the haploid-variant of Dipformer, and a CNN-based genomics model called DeepSEA. 

\section{Future Work}
For future works, we expect to focus on model explainability by using multiple analysis methods, such as analyzing the attention behaviour, analyzing the gradient saliency map, etc, to gather and verify insights from the model. Evaluation on larger scale dataset is also necessary to further demostrate the effectiveness of SNP2Vec. Additionally, adoption of SNP2Vec to other hereditary disorders and other complex genomics tasks is also an essential direction for future works.

\section*{Acknowledgement}

This work has been partially funded by School of Engineering PhD Fellowship Award, the Hong Kong University of Science and Technology and PF20-43679 Hong Kong PhD Fellowship Scheme, Research Grant Council, Hong Kong.



\bibliography{anthology,custom}
\bibliographystyle{acl_natbib}

\newpage

\appendix

\section{Mapping of SNP Tokens}
\label{app:snp-tokens-mapping}

\begin{CJK*}{UTF8}{gbsn}

\begin{table*}[!t]
\centering
\resizebox{1.0\linewidth}{!}{
\begin{tabular}{lr|lr|lr|lr|lr|lr}
\toprule
\multicolumn{12}{c}{\textbf{Mapping of SNP Tokens}} \\ \midrule
\texttt{A/A} & \texttt{A} & \texttt{DEL/A} & 腌 & \texttt{A/C} & 嗄 & \texttt{AI/C} & 爸 & \texttt{C/G} & 嚓 & \texttt{CI/G} & 懂 \\
\texttt{C/C} & \texttt{C} & \texttt{DEL/AI} & 拔 & \texttt{A/G} & 阿 & \texttt{AI/CI} & 比 & \texttt{C/N} & 拆 & \texttt{CI/GI} & 答 \\
\texttt{G/G} & \texttt{G} & \texttt{DEL/C} & 吃 & \texttt{A/N} & 呵 & \texttt{AI/G} & 八 & \texttt{C/T} & 礤 & \texttt{CI/N} & 达 \\
\texttt{N/N} & \texttt{N} & \texttt{DEL/CI} & 搭 & \texttt{A/T} & 锕 & \texttt{AI/GI} & 霸 & \texttt{C/CI} & 车 & \texttt{CI/NI} & 第 \\
\texttt{T/T} & \texttt{T} & \texttt{DEL/G} & 想 & \texttt{A/AI} & 吖 & \texttt{AI/N} & 巴 & \texttt{C/GI} & 床 & \texttt{CI/T} & 瘩 \\
\texttt{AI/AI} & \texttt{B} & \texttt{DEL/GI} & 香 & \texttt{A/CI} & 俺 & \texttt{AI/NI} & 逼 & \texttt{C/NI} & 穿 & \texttt{CI/TI} & 沓 \\
\texttt{CI/CI} & \texttt{D} & \texttt{DEL/N} & 学 & \texttt{A/GI} & 安 & \texttt{AI/T} & 把 & \texttt{C/TI} & 出 & \texttt{G/GI} & 高 \\
\texttt{GI/GI} & \texttt{H} & \texttt{DEL/NI} & 虚 & \texttt{A/NI} & 案 & \texttt{AI/TI} & 笔 & \texttt{GI/N} & \begin{CJK}{UTF8}{bkai}蝦\end{CJK} & \texttt{G/N} & 给 \\
\texttt{NI/NI} & \texttt{O} & \texttt{DEL/T} & 徐 & \texttt{A/TI} & 按 & \texttt{N/NI} & 讷 & \texttt{GI/NI} & 合 & \texttt{G/NI} & 股 \\
\texttt{TI/TI} & \texttt{U} & \texttt{DEL/TI} & 需 & \texttt{NI/T} & 喔 & \texttt{N/T} & 哪 & \texttt{GI/T} & 虾 & \texttt{G/T} & 个 \\
\texttt{DEL/DEL} & \texttt{X} & \texttt{T/TI} & 拓 & \texttt{NI/TI} & 侬 & \texttt{N/TI} & 娜 & \texttt{GI/TI} & 盒 & \texttt{G/TI} & 该 \\

\bottomrule
\end{tabular}
}
\caption{Mapping of SNP tokens into a single character representation.} 
\label{tab:snp-tokens-mapping}
\end{table*}

\end{CJK*}

As our resulting SNP tokens are represented as `X$_1$/X$_2$'', to reduce the size and facilitate more straightforward representation for the downstream process in the NLP pipeline, such as pre-processing, tokenization, and modeling, we map all SNP tokens into a single character representation. The mapping of the SNP tokens into a single character representation is shown in Table~\ref{tab:snp-tokens-mapping}. We use non-alphabetical characters as there are 66 SNP tokens in total, more than the available alphabetical characters, which consists of 52 characters (lower and upper case from `A' to `Z') in total. Also note that, all the SNP tokens related to the unkown token `N' except `N/N' (such as `A/N', `G/NI', `N/NI', `NI/NI', etc) are never been used since there is no actual SNP record corresponding to the unknown token `N'. The combinations of all `N' and `NI' tokens are listed on the table only for completion.


\section{Model Hyperparameters}
\label{app:model_hyperparam}

We develop two Linformer~\cite{wang2020linformer} models, i.e., Dipformer and Hapformer, which is pre-trained using our proposed SNP tokens and the original nucleotide tokens, respectively. The two models have the same hyperparameter settings resulting in an equal number of parameters. We list all the hyperparameters of our Dipformer and Hapformer models in Table~\ref{tab:model_hyperparam}.

\begin{table*}[t]
\begin{adjustbox}{width=\linewidth,totalheight={\textheight},keepaspectratio}
\begin{minipage}[!b]{0.3\linewidth}
    \centering
    \begin{tabular}{lc}
    \toprule
    \textbf{Hyperparams}          & \textbf{Value}  \\ \midrule
    \#layers & 6  \\
    dim & 512 \\
    k & 128 \\
    dropout & 0.1 \\
    num heads & 8 \\
    dim head & 64 \\ 
    num embeddings & 32000 \\
    single KV head & False \\
    shared KV & False \\ \bottomrule
    \end{tabular}
    \caption{Model Hyperparameters}
    \label{tab:model_hyperparam}
\end{minipage}
\hspace{3mm}
\begin{minipage}[!b]{0.35\linewidth}
    \centering
    \begin{tabular}{lc}
    \toprule
    \textbf{Hyperparams}          & \textbf{Value}  \\ \midrule
    batch size & 240 \\
    optimizer & AdamW\\
    learning rate & 1e-4 \\
    scheduler $\lambda$1 & 1 \\
    scheduler $\lambda$2 & 0.999991 \\
    \#steps & 200,000 \\
    warmup step & 1000 \\
    loss fn & Cross Entropy \\
    random seed & 0 \\ \bottomrule
    \end{tabular}
    \caption{Pre-Training Hyperparameters}
    \label{tab:pretrain_hyperparam}
\end{minipage}
\hspace{3mm}
\begin{minipage}[!b]{0.34\linewidth}
    \centering
    \begin{tabular}{lc}
    \toprule
    \textbf{Hyperparams}          & \textbf{Value}  \\ \midrule
    batch size & 16 \\
    optimizer & AdamW\\
    learning rate & [1e-4..1e-6] \\
    scheduler $\lambda$1 & 1 \\
    scheduler $\lambda$2 & 0.999991 \\
    \#epoch & 30 \\
    early stopping & 3 \\
    loss fn & Cross Entropy \\
    random seed & 0 \\ \bottomrule
    \end{tabular}
    \caption{Fine-Tuning Hyperparameters}
    \label{tab:finetuning_hyperparam}
\end{minipage}
\end{adjustbox}
\end{table*}

\section{Pre-Training Setup}
\label{app:pretrain_hyperparam}

During the pre-training phase, we build the BPE tokenizer with a vocab size of 32,000 for both the SNP tokens and nucleotide tokens datasets. We perform pre-training on both Dipformer and Hapformer models for 200,000 steps using masked language modeling with the cross entropy loss. During the pre-training, we apply a masking strategy similar to BERT~\cite{devlin2019bert} with a 15\% token replacement rate, where we replace with \texttt{[MASK]} 80\% of the time, replace with a random token 10\% of the time, and keep the token as is 10\% of the time. We run the pre-training using 5 units of 2080Ti GPUs and an Intel(R) Xeon(R) Silver 4210 CPU. We use the same hyperparameter settings for pre-training both the Dipformer and Hapformer models. The hyperparameters of our pre-training are shown in Table~\ref{tab:pretrain_hyperparam}.

\section{Fine-Tuning Setup}
\label{app:finetuning_hyperparam}

We fine-tune all models on Alzheimer's disease risk prediction on a Chinese cohort consisting of 624 subjects in total, 384 of which are labelled as Alzheimer's disease carriers (ADs) while 240 others are non-carriers (NCs). For predicting Alzheimer's disease, we append a \texttt{[CLS]} token as the prefix of the sequence. During the fine-tuning, we take the output of the \texttt{[CLS]} token and perform a linear transformation on it to get the disease risk prediction. We evaluate the performance of all models using accuracy, area underthe  ROC curve (AUROC), and area under the precision-recall curve (AUPRC). We show all the hyperparameters of the fine-tuning phase in Table~\ref{tab:finetuning_hyperparam}. We experiment with different learning rate for each model and find that the best setting is achieved when using a learning rate of 1e-4 for models that are not pre-trained (non-pre-trained Dipformer and DeepSEA) and a learning rate of 1e-5 for all pre-trained models (Dipformer and Hapformer).

\section{Detailed Results}
\label{app:results_detail}
\begin{figure*}[t]
    \centering
    \includegraphics[width=1.0\linewidth]{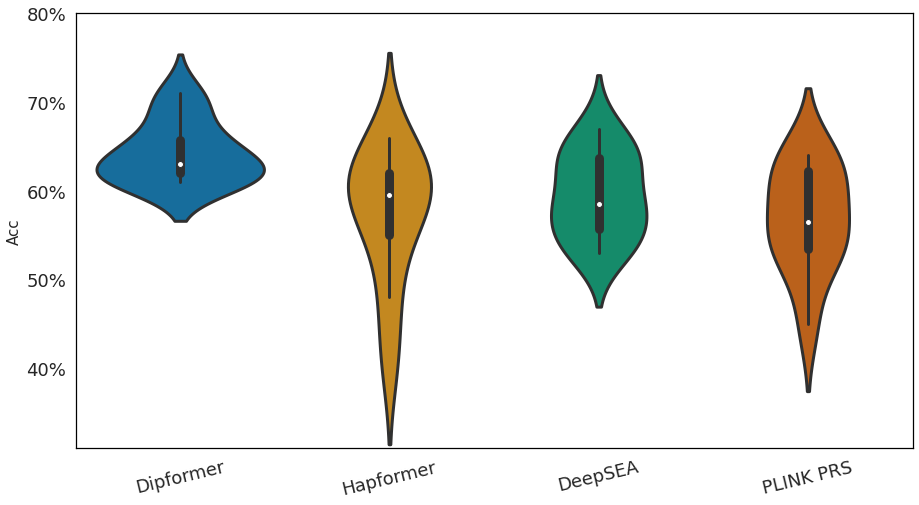}
    \captionof{figure}{10-folds accuracy performance of the best Dipformer, Hapformer, DeepSEA, and PLINK models on the Alzheimer's disease risk prediction.}
    \label{fig:acc-violin}
\end{figure*}

\begin{figure*}[t]
    \centering
    \includegraphics[width=1.0\linewidth]{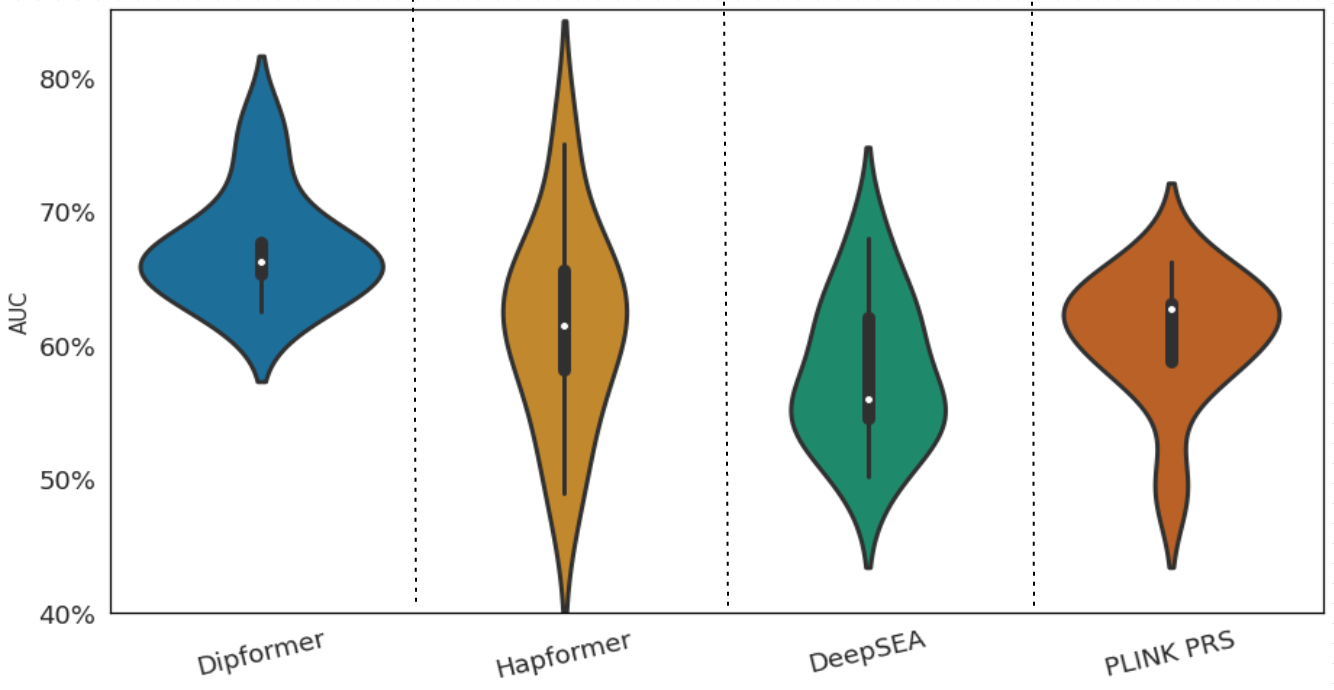}
    \captionof{figure}{10-folds AUROC performance of the best Dipformer, Hapformer, DeepSEA, and PLINK models on the Alzheimer's disease risk prediction.}
    \label{fig:auroc-violin}
\end{figure*}

\begin{figure*}[!t]
    \centering
    \includegraphics[width=1.0\linewidth]{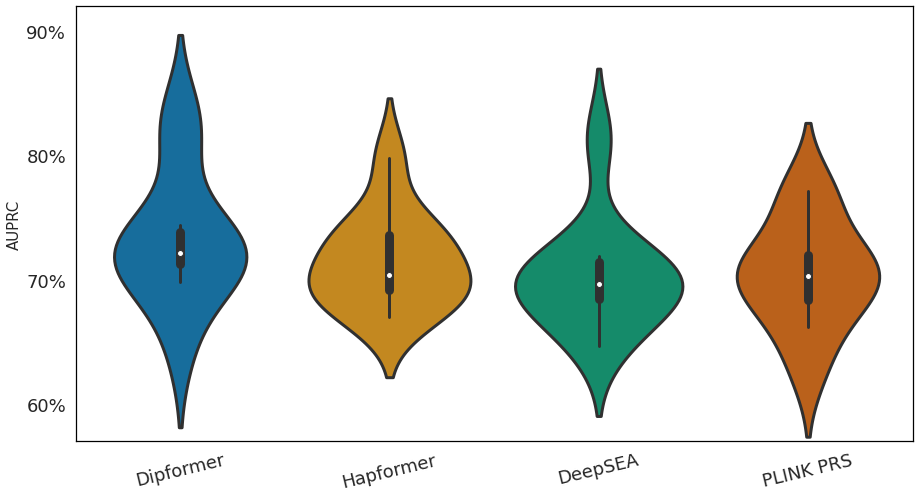}
    \captionof{figure}{10-folds AUPRC performance of the best Dipformer, Hapformer, DeepSEA, and PLINK models on the Alzheimer's disease risk prediction.}
    \label{fig:auprc-violin}
\end{figure*}

In this section, we show the distribution of the 10-fold results from our experiment in the Alzheimer's disease risk prediction task for all models (Dipformer, Hapformer, DeepSEA, and PLINK) on each evaluation metric.  Figure~\ref{fig:acc-violin} shows the distribution of the best 10-folds accuracy  performance on the Alzheimer's disease risk prediction task. Figure~\ref{fig:auroc-violin} shows the distribution of the best 10-folds AUROC  performance on the Alzheimer's disease risk prediction task. Figure~\ref{fig:auprc-violin} shows the distribution of the best 10-folds AUPRC performance on the Alzheimer's disease risk prediction task.


\end{document}